\documentclass[preprint]{vgtc}

\ifpdf
  \pdfoutput=1\relax                   
  \usepackage{graphicx}                
  \DeclareGraphicsExtensions{.pdf,.png,.jpg,.jpeg} 
\else
  \usepackage{graphicx}                
  \DeclareGraphicsExtensions{.eps}     
\fi%

\graphicspath{{figures/}{pictures/}{images/}{./}} 

\usepackage{microtype}                 
\PassOptionsToPackage{warn}{textcomp}  
\usepackage{textcomp}                  
\usepackage{mathptmx}                  
\usepackage{times}                     
\usepackage{cite}                      
\usepackage{tabu}                      
\usepackage{booktabs}                  
\usepackage{hyperref}

\preprinttext{}

\title{LossPlot: A Better Way to Visualize Loss Landscapes}

\author{Robert Bain\thanks{e-mail: bainro@oregonstate.edu}\\ %
        \scriptsize Oregon State University %
\and Mikhail Tokarev\thanks{e-mail: tokarevm@oregonstate.edu}\\ %
        \scriptsize Oregon State University %
\and Harsh Kothari\thanks{e-mail: kotharih@oregonstate.edu}\\ %
        \scriptsize Oregon State University %
\and Rahul Damineni\thanks{e-mail: daminens@oregonstate.edu}\\ %
        \scriptsize Oregon State University}

\abstract{
Investigations into the loss landscapes of deep neural networks are often laborious. This work documents our user-driven approach to create a platform for semi-automating this process. \textit{LossPlot} accepts data in the form of a csv, and allows multiple trained minimizers of the loss function to be manipulated in sync. Other features include a simple yet intuitive checkbox UI, summary statistics, and the ability to control clipping which other methods do not offer.}

\nocopyrightspace

\begin{document}
\maketitle

\begin{figure*}[tb]
  \centering
  \includegraphics[width=\linewidth]{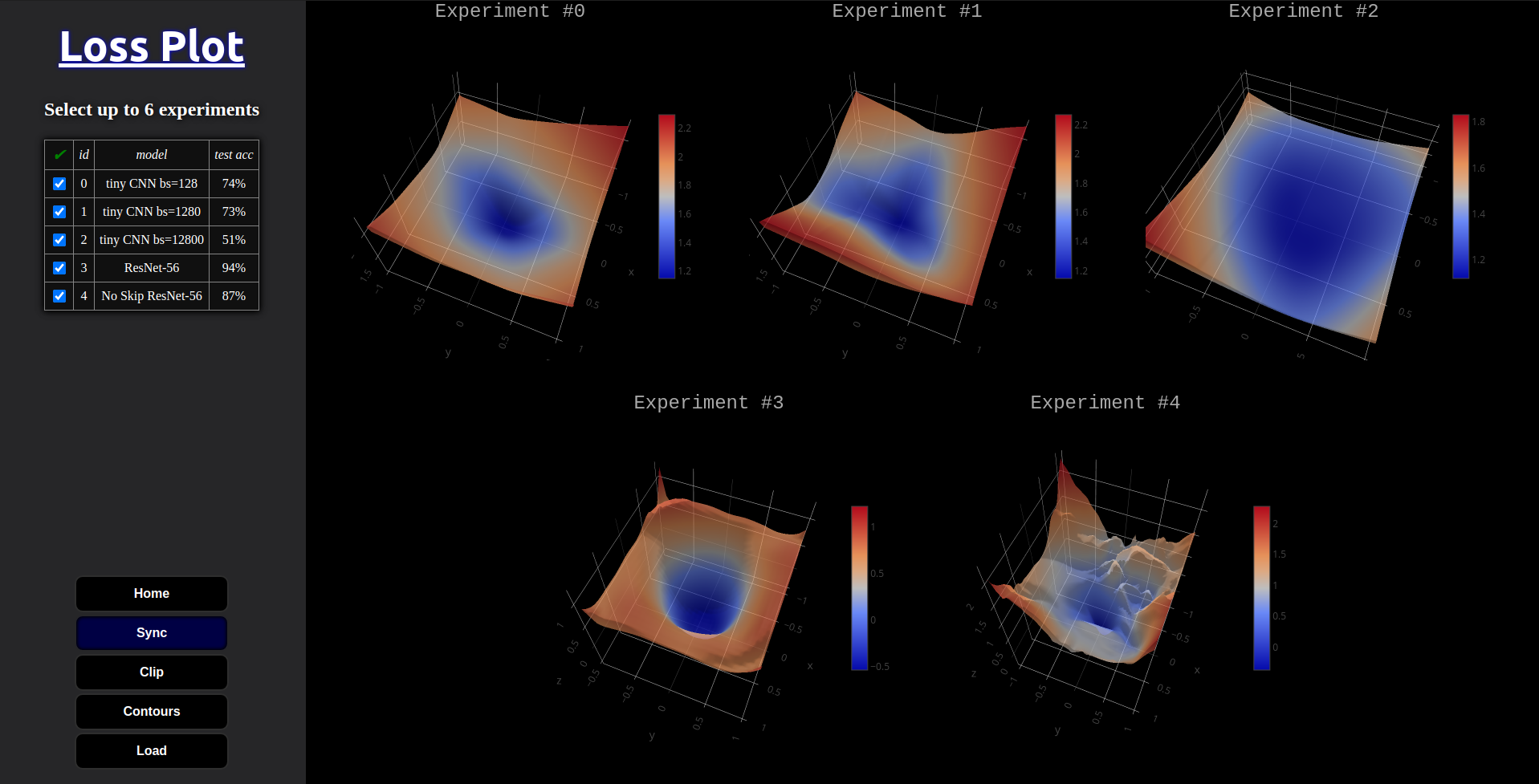}
  \caption{LossPlot Interface: The left side of the screen is the control panel in gray, and towards the right side of the screen is the plotting canvas. The plots are 3D surface plots enabling visualization of DNN's loss landscapes. Users can view data summary statistics in a tabular format at the top of the control panel. The data table has a simple checkbox UI that enables control over which of the up to six experiments are plotted. The buttons at the bottom of the control panel offer global controls over all individual plots.}
  \label{fig:banner}
\end{figure*}

\section{Introduction}
Deep Neural Networks (DNN) have gained growing attention in recent years by achieving SoTA performance in several task domains. Visualization tools aimed at distilling and debugging components of DNN models  contributed to this success by enabling researchers to understand contemporary algorithms' shortcomings and generate solutions that address them. TF Playground \cite{smilkov2017directmanipulation} is aimed at understanding overall dynamics and interplay of these components. TF Graph Visualizer \cite{8019861} focuses on model architecture. The optimisation of a DNN involves minimizing a high-dimensional non-convex loss function. This task is difficult in theory but as with other components of DNNs, intuition can be gained from visual aides. The trainability and complexity of DNNs is dependent on many factors including the chosen architecture, optimizer, initialization, learning rates, and other hyperparameters. Each of these factors effects on training and the resulting model's performance are often unclear. Visualizations of models’ weight spaces evaluated for a given task’s loss (i.e. loss landscapes) have been used to answer many interesting questions in the past \cite{1712.09913, 1412.6544, 1803.00885, 1910.03867}. Many questions remain unanswered though, including do different model minima have comparable local landscapes \cite{1803.00885}, is loss landscape smoothness a good heuristic for a model's generalization error, and what causes performance degradation when increasing DCNN's batch size beyond a certain amount \cite{1712.09913}? We introduce LossPlot to help answer some of these questions. LossPlot employs 3D Plotly.js surface plots to visualize 2D grid slices of DNNs’ loss landscapes.

All visualizations of the loss landscape were ultimately generated using the methods from \cite{1712.09913}. Their open-sourced pytorch code evaluates a 2d grid slice centered around a trained minimizer. There are many 2d grid slices to choose from given the dimensionality of modern DNNs. \cite{1712.09913} uses a dimensionality reduction technique, akin to UMAP \cite{mcinnes2020umap}, tSNE \cite{Maaten2008VisualizingDU}, and PCA \cite{sorzano2014survey}, to choose the slice of the weight space plotted for its loss. At a high level, the method begins with creating two random weight vectors by sampling a Gaussian distribution N(0, 1). Information about the learned weight solution is then incorporated into the weights in one of several ways. We use filter normalization for all of our experiments. This scales each corresponding filter in the random weight space direction by the learned filter’s Frobenius weight norm. Please refer to \cite{1712.09913} for a more detailed explanation.
 
LossPlot allows comparison of multiple model’s loss landscapes and even comparison between different loss landscapes projections. This tool will enable researchers to build better loss projections, and gain insight into the effects that training parameters and architectural choices have on said projections. Industry model builders and hobbyists can use this tool to gain trust in their model's ability to generalize to unseen data and ensure that the model of interest is complex enough for future needs. We outline a user case where we explore the empirical observations that increasing the mini-batch size too much will negatively affect model's performance. \cite{8019861} has hypothesized that this effect could in part be due to reduced noise in the larger mini-batches being attracted to exotic basin structures that allow for sharp minimizers to develop. We test this hypothesis in a similar manner as \cite{8019861}, but get different results.
 
The full test set has been used in the past to produce these loss surface plots, yet the distribution prior seems pretty well understood after only 100 validation examples. This novel observation leads to a 100x speedup over contemporary methods. Datasets lacking extreme outliers should benefit from this method. The authors of \cite{8019861} reported that it took hours to produce some of their plots using multi-gpu machines. They did not report on the resolution of their plots, but we were able to produce 6 60x60 plots in roughly 45 minutes. 
 
We want to enable further research into loss landscape dimensionality reduction techniques. These types of visualizations are getting better at capturing the complexity and trainability \cite{8019861, landscape} of DNN models. Visualizing these projections can give spatial intuition to both ML researchers and engineers, giving them the ability to debug their models through multiple rounds of experimentation and thus improving their trust in their models.  
 
Empirical intuition has been dominant in the field for the last decade of DNNs, with theory often lagging behind. These types of visualization can help give a handle to what is going on behind the scenes, enabling researchers to test hypotheses relating to proposed frameworks of explanation. I.e. they can theorize about the effects certain changes should have and this tool allows them to check them. On the flip side of that coin, those researchers need to be able to trust that the projections of the loss landscape capture the whole story, or at the least the information they care about. So to enable that trust to be built, our tool is not dependent on any single projection method and was built with the intention of being used to test future methods side by side with contemporary ones.
 
We have a wide set of target users. Essentially anyone who builds ML models could benefit from our tool. Researchers who want to do experiments to test their theoretical frameworks, engineers in industry who need to debug their models, or just hobbyists who might want some intuition as to how their decisions are affecting the underlying loss landscape and thus training via SGD and its many variants. 
 
There are two pre-existing tools that partially meet our users' needs. The first is an online browser tool \cite{telesense}. It meets only the first 2 outlined user needs: 1) It allows viewing a single model and 2) it allows interacting with that model. It cannot compare models, users cannot view custom data, there are only a handful of pre-calculated plots to choose from, and all but one of the UI elements are useless given our particular user needs.  
 
The other method \cite{tomgoldstein} is powerful, but has a high barrier to entry. It involves exporting your projection to a .vtk file, which is an unusual legacy format. You must then setup a Paraview server to view this .vtk file. The setup process is quite involved and has its own lengthy wiki style guide, with 10s of steps to do so. You must compile the source code, install it, set file system permissions, set networking permissions, etc. Paraview also has a very complicated UI, which while very powerful, has a steep learning curve. This is overkill for our users, as they would not even use 1/10th of the functionality that Paraview provides. 
 
Our tool is an in-browser tool that allows multiple models to be compared side-by-side. Each plot can be interacted with independently and quickly restored to the default position. Our tool does not cut off the corners of the grids, but instead gives users control over that ability using a "clipping" button. The clipping removes corners which can sometimes cause "extreme z-tails," which can drastically change the aspect ratio of the graphs when plotting, making side-by-side comparison difficult [figure!]. 
 
Our tool accepts a simple .csv file format and only requires 4 data column types: id, x, y, and loss. The x and y are the 2 directions in the weight space which define the 2d grid slice of the weight space being evaluated for its loss.  
 
We give users the ability to load their own experimental data through their local file system or through a URL. This enables quick sharing and makes the tool applicable to more than just curious enthusiasts.

\section{Design Goals}

\begin{itemize}
    \item \textbf{N1:} Visualize models’ loss landscape projections. The main need of our users was to gain insight into the loss landscape of a given model and task.
    \item \textbf{N2:} Interact with the projections (e.g. panning, zooming, rotating, etc). We aimed to give a general overview, but allow for details to be available quickly through the aforementioned interactions.
    \item \textbf{N3:} Compare multiple models. This feature is extremely important for DNN researchers. Often people who tune DNNs employ multiple rounds of debugging, iteration, and experimentation. That subset of our users needed to see their decisions effects on the underlying loss surface during model tuning.
    \item \textbf{N4:} Benchmark future loss landscape projection techniques. If our tool is to remain relevant, it must change with the times. Just as tSNE has been largely superseded by UMAP, perhaps the projection technique we currently rely on will be improved as well.
    \item \textbf{N5:} Upload their own data and share it with others. Our users found that pre-existing tools either do not enable viewing of custom data or make it very difficult to share by requiring the installation of new software.
\end{itemize}

\section{Proposed Approach}

\begin{figure*}[tb]
  \centering
  \includegraphics[width=\linewidth]{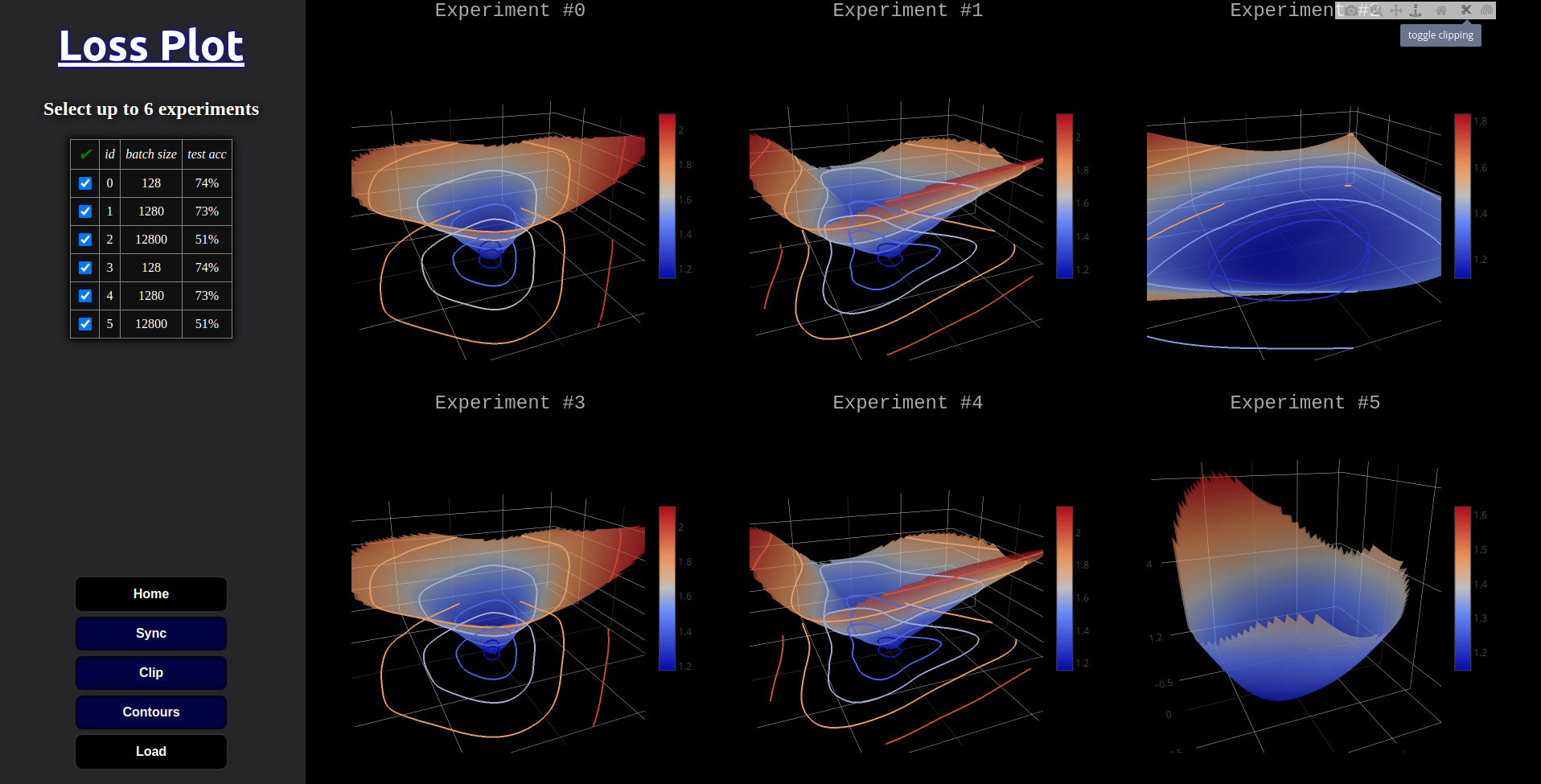}
  \caption{The results of varying batch size by orders of magnitude for a small DCNN trained on CIFAR10. The loss in accuracy is small when going from 128 to 1,280 and this might be explained by the less optimal loss surface. It is not round, but pinched in certain areas, which would lead to suboptimal gradient based updates, assuming that the found solution is a local minimizer. Each column of the plotting canvas shows the same data with minor differences. The two left-most columns only differ in the 2D slice of the weight space projected onto (i.e. different random direction vectors). The right-most column shows that the global controls in the control panel can also be individually controlled as well via an on hover "modebar." The bottom plot has had 2D contours turned off. The top plot has turned off radius clipping, which has a dramatic impact on the scaling.}
  \label{fig:lossplot}
\end{figure*}

LossPlot (\autoref{fig:lossplot}) is designed for three high-level functionalities that its antecedents did not get right: interactively exploring loss landscapes (N1, N2), comparing multiple landscapes effectively (N3, N4), and low effort sharing of the results (N5). To keep this tool as agnostic as possible to the details of the model, task, and training, we decoupled the plotting pipeline by simplifying the input to a .csv with pre-computed values. We expect the users to upload a .csv with projected weight dimensions (x, y), corresponding loss value (loss) along with an id column which is used for faster indexing. 

LossPlot has a control panel on the left and plotting area on the right. It has a data table viewer to control which experiments are plotted, also allowing individual features of each experiment to be quickly realized. At the bottom of the control panels are buttons. The “Contours” button helps in judging the smoothness and spread of individual contours by overlaying the corresponding 2d contours onto the 3d surface plots. The “Sync” button helps comparing multiple graphs by applying updates to all graphs simultaneously with the current user interaction. The “Load” button is used to enter data into the tool. The “Home” and “Clip” buttons improve usability of the tool by resetting graph view and clipping sometimes extreme corners of the 2d grid slices. 

Each of the aforementioned global controls which impact all plots have corresponding controls in the "modebar" which appears when hovering over the plots (see experiment \#3 in Figure 1). This enables users the most control over how each plot is displayed. Also in the modebar is a camera icon, which enables individual plots to be screen-captured and saved to the user's local file-system.

\section{Iteration}

\begin{itemize}
    \item Switched from Facebook’s Hiplot to Plotly.js’s parallel coordinate plot, as Hiplot is quite difficult to develop for as is evidence by the few people who have contributed to the github repo and the sparse documentation considering the many development pipelines they aim to offer.
 
    \item We removed the parallel coordinates plot, as brushing our data is non-intuitive and a simple html table with check boxes for each experiment row suffices. On top of that, visualizing 2D data on a spatial plot made more sense. With this extra room we were able to double the number of plots viewable at one time. 
 
    \item Added some global controls which affect all of the graphs at once, including clipping, 2d contours and home buttons. 
    
    \item Made the assumption that all experiments will be on the same size x-y plane, removing the need to brush x-y subsections of the data. Simplified the UI further.
 
    \item Our instructor provided aggregated feedback from our classmates about our final presentation. Several of our cohorts suggested that we sync up the interactions between our individual experiment plots. We felt this suggestion was an essential requirement for comparing multiple plots effectively. We have implemented this feature in the latest iteration of the tool. It is toggleable by a global control button below the data viewer panel with the rest of the buttons. Control over this is essential given that we do not assume that the 2d weight grids between models are of comparable scale. If this feature were always on it might give the impression of fair scaling comparison, when that is not necessarily a given, especially with custom user data. 
    
    \item We also removed a 4D plot that visualizes loss with respect to 3 input features (such as existence of skip-connections). We initially thought that such a visualization conveyed more information and it's a good idea to include it. However during demonstrations, it confused users. The variation in relative sizes of spatial points always required a walk-through during the demo and better visualized as a continuous surface. So, quality of visualization won over quantity of information.

\end{itemize}

\section{User Cases}
Here are the YouTube video links showcasing the following two user cases: \href{https://youtu.be/6n-_lA72lLg}{"Role of residuals"} and \href{https://youtu.be/2CD_mSzDJYQ}{"How test set size affects loss surface"}

Susan is creating a new DCNN architecture for an image classifier. She gets better results on her current model when adding skip connections. She reads about residuals and the identity function but is still confused as to why there is an improvement for her model. She samples the loss landscape in a 2d grid around her trained solutions (with and without skip connections) using pytorch and uploads them to our tool as a csv. She can visualize that resulting 3d surface in its entirety, and slice it many ways to gain further insights. She notices almost immediately that the skip connections lead to much smoother gradients in the loss landscape, allowing for better generalization error and trainability. 

Alexander is an ML researcher who is sceptical that the current dimensionality reduction techniques used to project down the full loss landscape of DNNs are ideal. He goes looking around the code and wonders how important it is to use all testing examples at each point in the weight space that the loss is computed. He creates several plots of varying number of validation examples and visualizes them in LossPlot. He can zoom in looking for regularities and mismatches between the different plots. He validates his hypothesis by gaining spatial intuition through our visualizations, without having to compute and compare mathematical characteristics of each plot's curvature.

\section{Feedback from Users}

We had an interview with the an ML researcher who most recently has been training DNNs for NLP tasks. Before the interview, we asked them to research on loss landscapes and their role in improving the model performance. We also hinted about ResNet's novelty and its important architecture difference with respect to others. Our aim was to have an informed user who could judge that smoothness of loss surface is attributed to residuals contributed by skip-connections. We hosted our tool online and let the user try it with a sample dataset. 

\begin{itemize}
    \item They discovered the control panel and realised expt. id field corresponds to various experiments with some help.
    \item Then they started toggling various experiments and explored loss landscapes. During this, they realised the prime difference between the two different class of experiments is skip-connections. 
    \item Almost immediately, they've realised that the landscapes with smooth surface had the skip-connections turned on. 
\end{itemize}

The user felt that the tool isn't end to end. They wanted an end-to-end solution that would let them spawn this visualization directly from the trained model instead of them generating the loss values and uploading them as a CSV. We felt that this was a legitimate ask because at the end, if the user has to manually tag each and every experiment they are performing, this tool is just as good as TensorBoard in book keeping aspect. We thought of a realistic solution that we discussed in "Discussion" section.

The user liked how intuitive the UI is and how he can nitpick details on loss surfaces using interactive tools. 

\section{Discussion}
This tool would be most useful when it is integrated into TensorBoard \cite{tensorflow}. Although TensorBoard includes utilities to monitor optimization trajectory by plotting scalars, it currently lacks the ability to visualize and compare loss landscapes. To enable the aforementioned utility, the loss and weight projection calculation logic should be integrated into a DNN library's model class (for example, PyTorch's nn.Module) and a method should be exposed to execute this pipeline and launch a simple visualization server. Alternatively, TensorBoard itself could be extended to include a "loss landscape" tab which will take a model, test dataset, and loss function to generate the visualizations. 
We attempted to extend the projections of the loss landscape from 2 spatial dimensions to 3. The motivation from this came from /cite{infoviz} in that most info visualization construction is first choosing the 3 spatial dimensions and then making do with what’s left. The 3d surface plots have redundant encoding of information in that the loss is both the Z axis and the color mapping. Adding another spatial dimension in the project allows for many more loss gradients to be visualized at once, in theory.

We wanted to investigate if the color mapping could be enough by itself to convey the loss. We used a 3d scatterplot, but quickly realized that color by itself was not going to convey the loss gradients as well as the 2d plots. The main issue was that you could not see the area of interest, the center of the 3d cubic slice of the weight space where the minimizer was. We tried inverting the marker size and softmax scaling the loss in an attempt to draw attention to the middle of the cube where the low loss values would ideally be. This worked to an extent but neither this or any of the other visualizations we created seemed as intuitive as the hill like surfaces which we could create with the loss encoded in both color and the z-axis.
 
Weight and Biases is coming up as one of the de facto model training and comparison communities. They jsonify many different supported data type formats and send them to Weight and Biases' servers to be displayed on the frontend. One of the supported data formats is Plotly.js plots. This would be an interesting avenue to explore since we have lowered the work to produce these loss landscape projections 100x. A GPU is still required to compute the loss, but this is going to required of essentially all current DNN training. A GPU is not required to visualize the resulting projections. We confirmed this by viewing LossPlot with 6 surface plots rendered on a 5 year old dual-core laptop with integrated graphics was and it still runs very well.

\section{Contribution Statement}
All members participated in testing the multiple demos for bugs.
 
\begin{itemize}
    \item Robert came up with the main topic and led this work. He did the programming, created the datasets, setup the hosting solutions, and prepared the majority of slide show presentation materials.
    
    \item Mikhail helped brainstorm potential themes at the outset of the project. He acquired multiple user’s feedback, assisted with the final report, and coordinated each member’s contributions to the paper prototype. 
    
    \item Harsh worked on the report, identified user needs, and created our paper prototype. 
    
    \item Rahul worked on finding user needs and contributed ideas to the team meetings. He also worked on the final report.
    
\end{itemize}

\acknowledgments{
The authors wish to thank Prof. Minsuk Kahng and Dayeon Oh.}

\bibliographystyle{abbrv-doi-hyperref}
\bibliography{references}

\begin{thebibliography}{10}

\bibitem{landscape}
\href{https://losslandscape.com/}{Loss landscape: A.i deep learning
  explorations of morphology \& dynamics}.

\bibitem{tensorflow}
M.~Abadi, A.~Agarwal, P.~Barham, E.~Brevdo, Z.~Chen, C.~Citro, G.~S. Corrado,
  A.~Davis, J.~Dean, M.~Devin, S.~Ghemawat, I.~Goodfellow, A.~Harp, G.~Irving,
  M.~Isard, Y.~Jia, R.~Jozefowicz, L.~Kaiser, M.~Kudlur, J.~Levenberg, D.~Mane,
  R.~Monga, S.~Moore, D.~Murray, C.~Olah, M.~Schuster, J.~Shlens, B.~Steiner,
  I.~Sutskever, K.~Talwar, P.~Tucker, V.~Vanhoucke, V.~Vasudevan, F.~Viegas,
  O.~Vinyals, P.~Warden, M.~Wattenberg, M.~Wicke, Y.~Yu, and X.~Zheng.
\newblock Tensorflow: Large-scale machine learning on heterogeneous distributed
  systems, 2016.

\bibitem{1803.00885}
F.~Draxler, K.~Veschgini, M.~Salmhofer, and F.~A. Hamprecht.
\newblock Essentially no barriers in neural network energy landscape.
\newblock 2018.

\bibitem{tomgoldstein}
\href{https://github.com/tomgoldstein/loss-landscape\#visualizing-3d-loss-surface}{T.~{Goldstein}}.
\newblock
  \href{https://github.com/tomgoldstein/loss-landscape\#visualizing-3d-loss-surface}{https://github.com/tomgoldstein/loss-landscape\#visualizing-3d-loss-surface}.

\bibitem{1412.6544}
I.~J. Goodfellow, O.~Vinyals, and A.~M. Saxe.
\newblock Qualitatively characterizing neural network optimization problems,
  2014.

\bibitem{1712.09913}
H.~Li, Z.~Xu, G.~Taylor, C.~Studer, and T.~Goldstein.
\newblock Visualizing the loss landscape of neural nets, 2017.

\bibitem{Maaten2008VisualizingDU}
L.~V.~D. Maaten and G.~E. Hinton.
\newblock Visualizing data using t-sne.
\newblock {\em Journal of Machine Learning Research}, 9:2579--2605, 2008.

\bibitem{mcinnes2020umap}
L.~McInnes, J.~Healy, and J.~Melville.
\newblock Umap: Uniform manifold approximation and projection for dimension
  reduction, 2020.

\bibitem{telesense}
\href{https://www.telesens.co/loss-landscape-viz/viewer.html}{A.~{Mohan}}.
\newblock \href{https://www.telesens.co/loss-landscape-viz/viewer.html}{Loss
  visualization}.

\bibitem{1910.03867}
I.~Skorokhodov and M.~Burtsev.
\newblock Loss landscape sightseeing with multi-point optimization, 2019.

\bibitem{smilkov2017directmanipulation}
D.~Smilkov, S.~Carter, D.~Sculley, F.~B. Viégas, and M.~Wattenberg.
\newblock Direct-manipulation visualization of deep networks, 2017.

\bibitem{sorzano2014survey}
C.~O.~S. Sorzano, J.~Vargas, and A.~P. Montano.
\newblock A survey of dimensionality reduction techniques, 2014.

\bibitem{8019861}
\href{https://doi.org/10.1109/TVCG.2017.2744878}{K.~{Wongsuphasawat},
  D.~{Smilkov}, J.~{Wexler}, J.~{Wilson}, D.~{Mané}, D.~{Fritz},
  D.~{Krishnan}, F.~B. {Viégas}, and M.~{Wattenberg}}.
\newblock \href{https://doi.org/10.1109/TVCG.2017.2744878}{Visualizing dataflow
  graphs of deep learning models in tensorflow}.
\newblock \href{https://doi.org/10.1109/TVCG.2017.2744878}{{\em IEEE
  Transactions on Visualization and Computer Graphics}},
  \href{https://doi.org/10.1109/TVCG.2017.2744878}{24(1):1--12},
  \href{https://doi.org/10.1109/TVCG.2017.2744878}{2018}.
  \href{https://doi.org/10.1109/TVCG.2017.2744878}
{doi: {{%
10\hspace{.1pt}\discretionary{.}{%
}{.}\hspace{.4pt}1109\discretionary{/}{%
}{/}TVCG\hspace{.1pt}\discretionary{.}{%
}{.}\hspace{.4pt}2017\hspace{.1pt}\discretionary{.}{%
}{.}\hspace{.4pt}2744878}}}


\end{thebibliography}
\end{document}